\documentclass[conference]{IEEEtran}
\IEEEoverridecommandlockouts
\usepackage{cite}
\usepackage{amsmath, amssymb, amsfonts}
\usepackage{algorithmic}
\usepackage{graphicx}
\usepackage{textcomp}
\usepackage{xcolor}
\usepackage{multirow}
\usepackage{amssymb}
\usepackage{etoolbox}
\usepackage{booktabs}
\usepackage{bm}
\usepackage{hyperref}
\usepackage[american]{babel}
\usepackage{microtype}
\newcommand{\RNum}[1]{\uppercase\expandafter{\romannumeral #1\relax}}

\def\BibTeX{{\rm B\kern-.05em{\sc i\kern-.025em b}\kern-.08em
 T\kern-.1667em\lower.7ex\hbox{E}\kern-.125emX}}
\begin{document}

\title{Visual-aware Attention Dual-stream Decoder\\for Video Captioning
\thanks{$\ast$ Corresponding author.}} %This work was supported in part by the Fundamental Research Funds for the Central Universities of China under Grant 191010001 and in part by the Hubei Key Laboratory of Transportation Internet of Things under Grant 2020\RNum{3}026GX.}}

\author{\IEEEauthorblockN{Zhixin~Sun$^1$, Xian~Zhong$^{1,2,\ast}$, Shuqin~Chen$^1$, Lin~Li$^{1,2}$, and~Luo~Zhong$^{1,2}$}
\IEEEauthorblockA{$^1$ \textit{School of Computer and Artificial Intelligence}, \textit{Wuhan University of Technology}, Wuhan, China\\
$^2$ \textit{Hubei Key Laboratory of Transportation Internet of Things}, \textit{Wuhan University of Technology}, Wuhan, China\\
$^3$ \textit{ZhongQianLiYuan Engineering Consulting Co., Ltd}, Wuhan, China\\
\{sunzx\_jdi, zhongx, csqcwx0801, cathylilin, zhongluo\}@whut.edu.cn}
%\and
%\IEEEauthorblockN{2\textsuperscript{nd} Xian Zhong}
%\IEEEauthorblockA{\textit{Hubei Key Laboratory of Transportation Internet of Things}\\
%\textit{School of Computer Science and Technology}\\
%\textit{Wuhan University of Technology}\\
%Wuhan, China\\
%zhongx@whut.edu.cn}
%\and
%\IEEEauthorblockN{3\textsuperscript{rd} Shuqin Chen}
%\IEEEauthorblockA{\textit{School of Computer Science and Technology}\\
%\textit{Wuhan University of Technology}\\
%Wuhan, China\\
%csqcwx0801@whut.edu.cn}
%\and
%\IEEEauthorblockN{4\textsuperscript{th} Lin Li}
%\IEEEauthorblockA{\textit{Hubei Key Laboratory of Transportation Internet of Things}\\
%\textit{School of Computer Science and Technology}\\
%\textit{Wuhan University of Technology}\\
%Wuhan, China\\
%cathylilin@whut.edu.cn}
% \and
% \IEEEauthorblockA{\textit{dept. name of organization (of Aff.)}\\
% City, Country\\
% \and
% \IEEEauthorblockA{\textit{dept. name of organization (of Aff.)}\\
% City, Country\\
}

\maketitle

\begin{abstract}

Video captioning is a challenging task that captures different visual parts and describes them in sentences, for it requires visual and linguistic coherence. 
The attention mechanism in the current video captioning method learns to assign weight to each frame, promoting the decoder dynamically. This may not explicitly model the correlation and the temporal coherence of the visual features extracted in the sequence frames.
To generate semantically coherent sentences, we propose a new Visual-aware Attention (VA) model, which concatenates dynamic changes of temporal sequence frames with the words at the previous moment, as the input of attention mechanism to extract sequence features.
In addition, the prevalent approaches widely use the teacher-forcing (TF) learning during training, where the next token is generated conditioned on the previous ground-truth tokens.
The semantic information in the previously generated tokens is lost. Therefore, we design a self-forcing (SF) stream that takes the semantic information in the probability distribution of the previous token as input to enhance the current token.
The Dual-stream Decoder (DD) architecture unifies the TF and SF streams, generating sentences to promote the annotated captioning for both streams.
%We train the whole network in an end-to-end manner by mixed training learning.
Meanwhile, with the Dual-stream Decoder utilized, the exposure bias problem is alleviated, caused by the discrepancy between the training and testing in the TF learning.
The effectiveness of the proposed Visual-aware Attention Dual-stream Decoder (VADD) is demonstrated through the result of experimental studies on Microsoft video description (\textbf{MSVD}) corpus and MSR-Video to text (\textbf{MSR-VTT}) datasets.
\end{abstract}

\begin{IEEEkeywords}
Video Captioning, Visual-aware Attention, Dual-stream Decoder, Mixed training learning, Exposure bias
\end{IEEEkeywords}

\section{\textbf{Introduction}}
Video Captioning is the task of generating a meaningful natural sentence for a given video. This task combines video understanding and natural language processing methods to generate informative and fluent captioning. Applications \textit{e.g.} video understanding, video summarization, and human-robot interaction depend on video captioning.

Recent works~\cite{DBLP:conf/aaai/FangZJZWZF19, DBLP:conf/ijcai/JinHCLZ20, DBLP:conf/ictai/LianLWH20, DBLP:conf/ictai/ZouLZZ20, DBLP:conf/cvpr/WuLCJL18, DBLP:conf/mm/LiuRY18, DBLP:conf/cvpr/PeiZWKST19, DBLP:conf/mm/ZhuJ19, DBLP:journals/npl/ChenZLLGZ20, DBLP:conf/cvpr/PanCHLGAN20} are based on encoder-decoder framework, where convolution neural network (CNNs) extracts vectorial representation from the video and recurrent neural networks (RNNs) decodes those representations into natural language sentences. Even though some recent works have explored precise and effective visual information for text generation, the situation in video captioning is still challenging because: 1) Unlike image captioning, which merely understands static content in a single image, video captioning requires a model to obtain the coherence of the consecutive frames; 2) To generate semantically rich and accurate sentences, it should take the polysemy and synonyms as considered.

% The attention module can provide precise and effective visual information for the text generation. 
% Most existing attention-based methods on visual captioning focus on the current word and visual information in one time step and generate the next word, without considering the visual and linguistic coherence. 
Inspired by the development of neural machine translation, the attention mechanism~\cite{DBLP:conf/aaai/ChenJ19, DBLP:conf/wacv/Cherian0HM20, DBLP:conf/iccv/HuangWCW19, DBLP:conf/ijcai/JinHCLZ20, DBLP:conf/aaai/FangZJZWZF19} has been widely used in current encoder-decoder frameworks for visual captioning, which guides the decoding process by generating a weighted average over the extracted feature vectors for each time step.
\cite{DBLP:conf/cvpr/ZhangP19a, DBLP:conf/mm/HuCZW19} apply a spatial/temporal attention mechanism to fuse object features. Although the attention module can provide precise and effective visual information for text generation, it ignores the visual relevance between adjacent words.
\cite{DBLP:conf/cvpr/QinDZL19} feeds the previous attention vector into the attention module to alleviate this problem. However, it still ignores the dynamic transmission and interrelationship of the visual features extracted by the attention mechanism for each frame.
\begin{figure}
	 \centering
	 \includegraphics[width = \columnwidth]{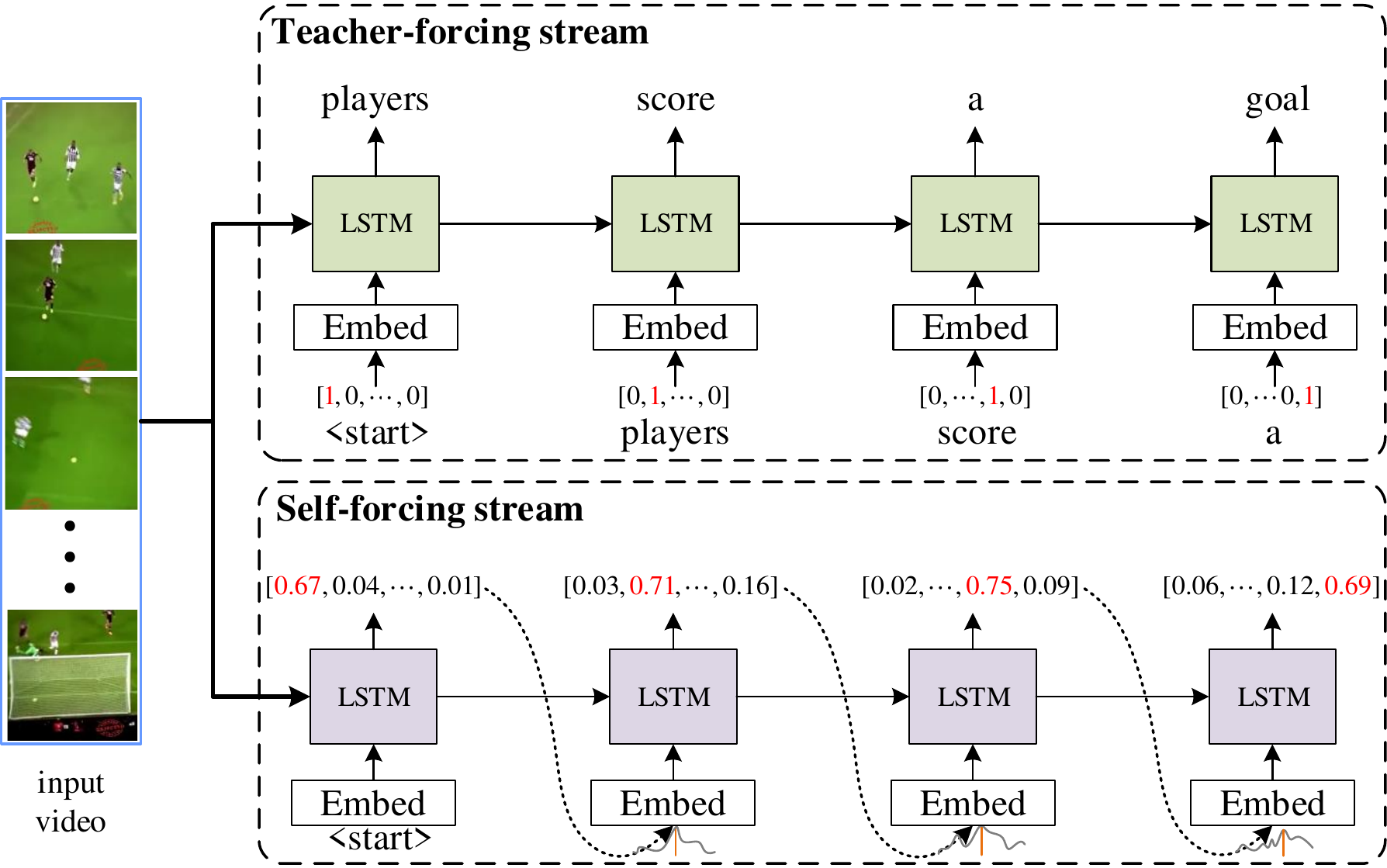}
	 \caption{\small The illustrate of our Dual-stream Decoder (DD) model, the teacher-forcing (TF) stream generates the predicted token based on the ground-truth token, while the self-forcing (SF) stream generates the current token based on the probability distribution of the previous token.}
	 \label{fig:pri}
	 %\vspace{-3mm}
\end{figure} 

For sentence generation, the teacher-forcing (TF) learning is used to generate the current token based on the previous ground-truth tokens in visual captioning~\cite{DBLP:conf/ijcai/TanLWZ20, DBLP:conf/iccv/VenugopalanRDMD15}. This causes the discrepancy between training and testing and the loss of semantic information in the probability distribution over the vocabulary. To address the exposure bias problem, some recent methods~\cite{DBLP:conf/emnlp/PasunuruB17, DBLP:conf/iccv/Wang00JWL19} directly optimize sentence-level task-based metrics (as rewards), using policy gradient and mixed-loss methods for reinforcement learning.
However, reinforcement learning solutions usually suffer from slow and unstable training due to the high variance of reinforce-based policy gradients. Zhang \textit{et al.}~\cite{DBLP:conf/ijcai/0009FL20} sample context words from the ground-truth sequences and the predicted sequence during training to alleviate the exposure bias issue. Nevertheless, the semantic information in the probability distribution of previously generated tokens is abandoned in these methods, which can not solve the impact of the ambiguity of the current word on the generation of the next word. 
 
For instance, in the predicted sentence of Figure~\ref{fig:pri}, the token “score” has two semantics. It can be used as a verb to indicate “goal scored” or as a noun “test score”. We take the output probability information (\textit{e.g.} score-60\%, goal-30\%, \textit{etc}.) of the previous tokens directly as the input of the current expression, retaining its most complete semantic information. 
% We use the probability distribution of previous tokens to guide the generation of current token.
% according to statistics of word frequency in caption corpus, exposure bias is caused by the discrepancy between training and testing. 

To solve the aforementioned problems, we propose Visual-aware Attention Dual-stream Decoder (VADD) for Video Captioning. Take a video input as an example, we first extract its vectorial representation with a pre-trained 2D/3D deep network and employ a bidirectional long short-term memory network (Bi-LSTM) to fuse sequences features. Visual-aware attention mechanism is used to capture the coherence between adjacent video frames. Parallel dual streams decoder, \textit{i.e.} TF stream and self-forcing (SF) stream, generate sentences to promote the annotated captions simultaneously. SF stream is designed to take all the probability distribution of previously generated tokens into the model. The mixed training loss is used to generate the sentence which unifies TF learning and SF learning. 

% Inspired by attention mechanism and Neural Machine Translation, we utilize visual-aware attention, teacher forcing learning and student recommend learning to generate more fluent sentence.
 
% We first extract its frame/clip representations with pretained 2D/3D deep network (e.g. ResNet). In the decoder stage, we use a two-stream decoder with parameter shared. The visual-aware attention mechanism is used to select the fused visual feature dynamically. The visual features extracted by the attention mechanism are tracked by an LSTM for each frame, the previous hidden state of LSTM2 and LSTM3 are concatenate as the input of attention mechanism for the next token generation. The teacher-forcing learning stream (TF) and the self-forcing learning stream (SF) are used to generate the probability distribution over the words in training simultaneously. 

% and we minimize the KL divergence between probability distribution of the TS and SR to transfer knowledge from TS to SR. 
% The previously probability distribution are conditioned to generate the next token in the SR, which makes the next token perceive richer semantic information of the previous token, and it is also the essential difference between the TS stream and the SF stream.

The main contributions of this paper are summarized threefold:
\begin{itemize}
	 \item We introduce a new visual-aware attention mechanism, an extension of the conventional attention mechanism, to model the dynamic transmission and relationship of the visual features extracted.
 
	 \item The dual-stream decoder is proposed for linguistic knowledge embedding from all previous tokens. The TF stream and the SF stream are used to generate sentences simultaneously during training. 
 
	 \item Experiments on commonly-used datasets for video captioning illustrate the effectiveness of our proposed VADD approach.
\end{itemize}

\begin{figure*}
	 \centering
	 \includegraphics[width = \textwidth]{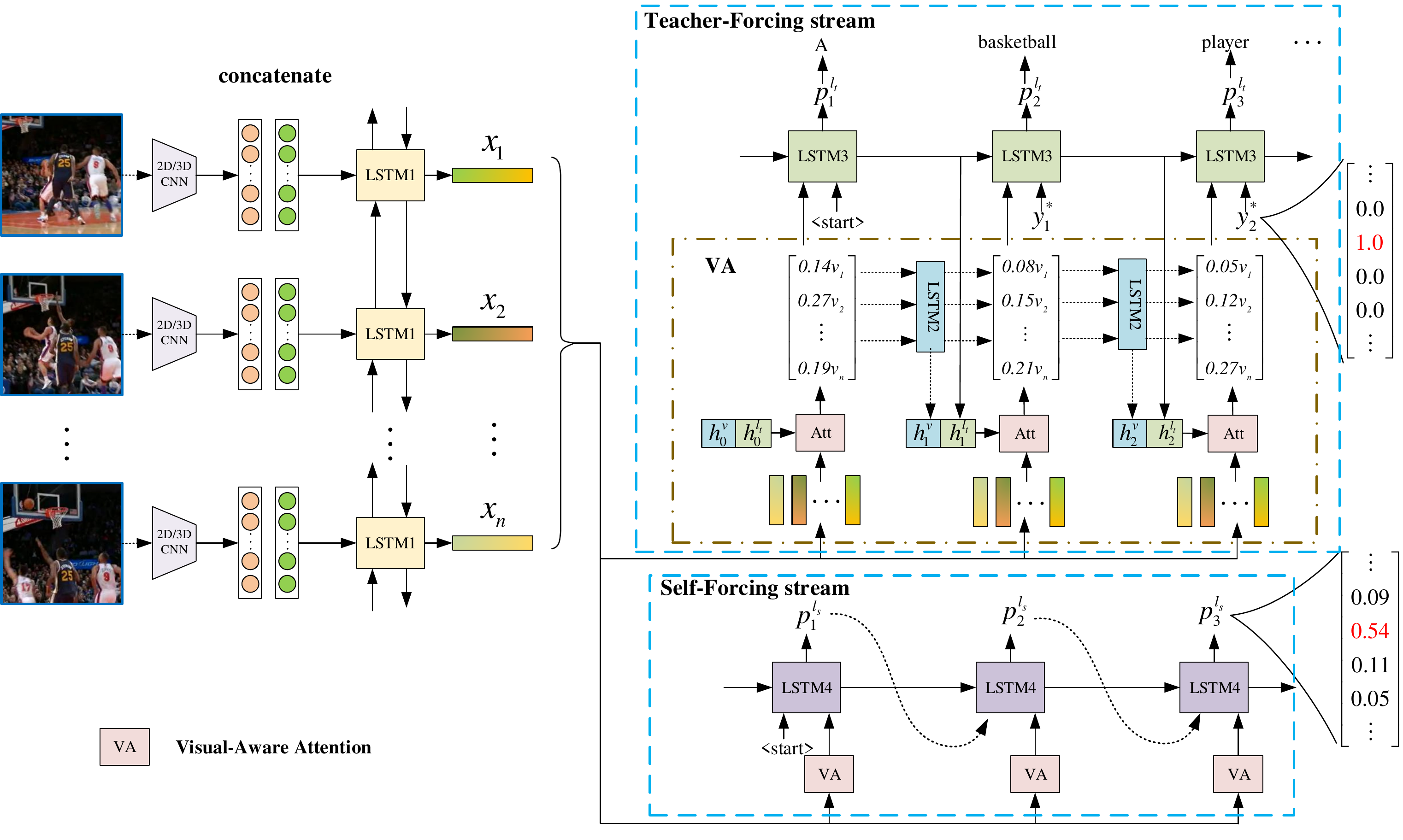}
 	 \caption{\small An overview of our framework with VADD. In the encoder stage, the 2D and 3D features are extracted for each frame, and a Bi-LSTM (LSTM1) is used to fuse the feature across time. In the decoder stage, we use the dual-stream decoder to take advantage of the semantic information in the probability distribution of the previous token. The visual-aware attention mechanism is used to select the fused visual feature dynamically. The visual features extracted by the attention mechanism are tracked by an LSTM for each frame, the previous hidden state of LSTM2 and LSTM3 are concatenated as the input of the attention mechanism for the next token generation. The TF stream and the SF stream are used to generate the probability distribution over the words in training simultaneously. The previous probability distribution is conditioned to generate the next token in the SF, which makes the next token perceive richer semantic information of the previous token, and it is also the essential difference between the TF stream and the SF stream.}
	 \label{fig:framework}
\end{figure*}

\section{\textbf{Related Work}}
 % \subsection{Video Captioning}
Visual captioning is a multi-modal problem that involves extracting key information from vision and describing it with natural language. With the great success of artificial intelligence~\cite{jiang2020decomposition, huang2020dotscn}, recent researches on visual captioning~\cite{DBLP:conf/ijcai/TanLWZ20, DBLP:conf/cvpr/ZhengWT20} adopt sequence-learning based methods in encoder-decoder structure. Visual captioning includes image captioning and video captioning, both of which use attention mechanisms to extract visual features and employ TF learning training methods to generate sentences. In the field of visual captioning, the attention mechanism is first introduced into image captioning~\cite{DBLP:conf/icml/XuBKCCSZB15} and achieves outstanding results. In recent years, the Huang \textit{et al.}~\cite{DBLP:conf/iccv/HuangWCW19} and Pan \textit{et al.}~\cite{DBLP:conf/cvpr/PanYLM20} improve the performance of the attention mechanism by further strengthening the interaction between the linguistic and visual features. Compared with the attention mechanism in image captioning, the attention mechanism in video captioning needs to consider the sequential coherence of visual features. Zhang \textit{et al.}~\cite{DBLP:conf/cvpr/ZhangP19a} and Cherian \textit{et al.}~\cite{DBLP:conf/wacv/Cherian0HM20} utilize a hierarchical attention mechanism to extract spatio-temporal visual features. Hu \textit{et al.}~\cite{DBLP:conf/mm/HuCZW19} and Fang \textit{et al.}~\cite{DBLP:conf/aaai/FangZJZWZF19} develop a coarse-to-fine attention mechanism to take both the global and local features. Taking the visual coherent into considering, Qin \textit{et al.}~\cite{DBLP:conf/cvpr/QinDZL19} feed the previous attention vector into the attention module. However, it still ignores the dynamic relationship transfer of the extracted visual features.

In the process of sentence generation, teacher-forcing (TF) learning which takes the previous ground-truth tokens as input, is widely used for captioning. TF will cause exposure bias problems, and also lose the semantic information in the probability distribution. Reinforcement learning is applied to optimize sentence-level task-based metrics using policy gradient and mixed-loss methods~\cite{DBLP:conf/emnlp/PasunuruB17, DBLP:conf/iccv/Wang00JWL19}. Zhang \textit{et al.}~\cite{DBLP:conf/ijcai/0009FL20} sample context words from the ground-truth sequences and the predicted sequence to alleviate the exposure bias issue, without the slow and unstable training problem suffered in reinforcement learning. Some reconstruction network~\cite{DBLP:conf/cvpr/Wang00018, DBLP:conf/cvpr/Chen0JY018} can also indirectly alleviate the exposure bias problem. Anyway, the rich semantic information in the probability distribution of the generated tokens is directly discarded without further exploration.

To solve the two problems mentioned above, we respectively proposed visual-aware attention mechanism and dual-stream decoder. The visual-aware attention mechanism employs a shared LSTM to construct the correlation and temporal coherence of the visual features extracted in the sequence frame. In dual-stream decoder, the SF stream is designed to take advantage of the semantic information of previous tokens. The SF stream is combined with the TF stream in a certain proportion to generate the final sentence.

\section{\textbf{Approach}}
% 还是分encoder, attention, decoder三部分讲比较好, 等下加一个encoder
In this section, we present the proposed video captioning approach based on visual-aware attention mechanism and dual-stream decoder in detail. The overall framework of the proposed VADD is illustrated in Figure~\ref{fig:framework}, where the input is the sequence of video frames $ \{v_{1}, v_{2}, \ldots, v_{n}\}$, and the output is the sequence of words $ \{y_{1}^{\ast}, y_{2}^{\ast}, \ldots, y_{m}^{\ast}\}$. For each video/sentence pair, we employ the pre-trained 2D/3D CNNs to extract appearance feature $\{v_{a_{1}}, v_{a_{2}}, \ldots, v_{a_{n}}\}$ and motion feature $\{v_{m_{1}}, v_{m_{2}}, \ldots, v_{m_{n}}\}$, where $n$ is the number of sampled frames. We first introduce our proposed video captioning framework in Subsections~\ref{sec:encoder}, \ref{sec:va}, and \ref{sec:dual-stream}, then describe the training and inference process in Subsection~\ref{sec:mtl}.

 % Then we concatenate $v_{a}$ and $v_{m}$ as the input of Bi-LSTM (bidirectional Long Short-Term Memory) to fuse the two kind of features across time, and we denote it as $\left (x_{1}, \ldots, x_{n}\right)$. 
\subsection{\textbf{Encoder}}
\label{sec:encoder}
% A Bidirectional Long Short-Term Memory network (Bi-LSTM) is used to encode the extracted video frames $\{v_{a_{1}}, v_{a_{2}}, \cdots, v_{a_{n}} \}$ and $\{v_{m_{1}}, v_{m_{2}}, \cdots, v_{m_{n}} \}$. 
% the annotation of $x_i$ is $h_i = \left[\overrightarrow{h}_i ; \overleftarrow{h}_i\right]$.
We concatenate $v_{a}$ and $v_{m}$ as the input of Bi-LSTM to fuse the two kinds of features across time, formally:
\begin{equation}
	 \overrightarrow{\bm{h}}_i\ = \ \overrightarrow{\text{LSTM1}} (v_{m_i} \circ v_{a_i}, \overrightarrow{\bm{h}}_{i-1})
\end{equation}
\begin{equation}
	 \overleftarrow{\bm{h}}_i\ = \ \overleftarrow{\text{LSTM1}} (v_{m_i} \circ v_{a_i}, \overleftarrow{\bm{h}}_{i-1})
\end{equation}
where $\overrightarrow{\text{LSTM1}}$ and $\overleftarrow{\text{LSTM1}}$ are the forward LSTM and the reverse LSTM of the bidirectional LSTM1 respectively, $\overrightarrow{\bm{h}}_i$ and $\overleftarrow{\bm{h}}_i$ are the output hidden states, and $\circ$ denotes concatenate operation. To obtain the encoded feature of each frame, the output forward hidden state $\overrightarrow{h}_i$ and the reverse hidden state $\overleftarrow{\bm{h}}_i$ are concatenated. In this way, the encoded feature of each frame summarizes the features of both the preceding frames and the following frames. We denote the outputs of the encoder as $\bm{V} = \{x_{1}, x_{2}, \ldots, x_{n}\}$, where $x_{n} = [\overrightarrow{\bm{h}}_i; \overleftarrow{\bm{h}}_i]$.

\begin{figure}
	 \centering
	 \includegraphics[width = 0.8\columnwidth]{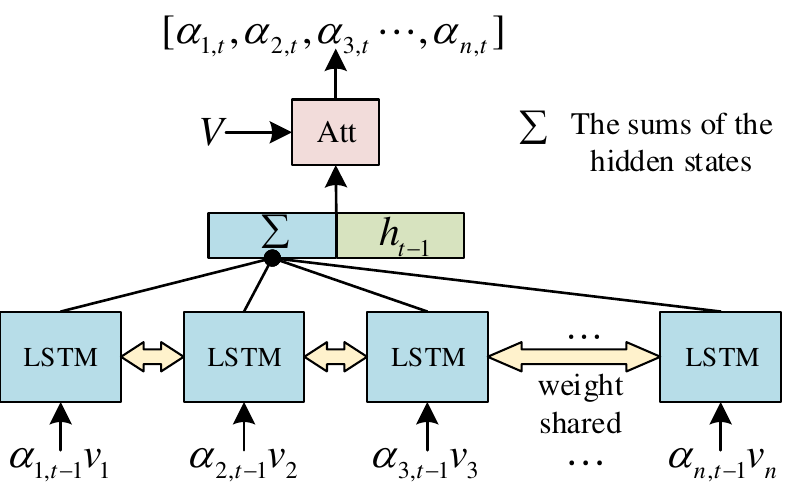}
	 \caption{\small The Visual-aware Attention (VA) module. We utilize a shared LSTM to construct the temporal coherence and dynamic relationship of the visual features extracted in each frame, then we sum these features and concatenate it with the previous hidden state ($\bm{h}_{t-1}^{l_t}$ in~\eqref{equ:tf-hidden} and $\bm{h}_{t-1}^{l_s}$ in~\eqref{equ:sf-hidden}) to extract the visual feature for the current token.}
	 \label{fig:vlstm}
	 %\vspace{-3mm}
\end{figure}

\subsection{\textbf{Visual-aware Attention Mechanism}}
\label{sec:va}
Figure~\ref{fig:vlstm} shows the visual-aware attention module. Given a video sequence feature vectors $\bm{V}$ and the previous hidden state $\bm{h}_{t-1}$, conventional attention module employs the attention function $f_{att}$ to calculate a weighted average vector $c_t$ as:
\begin{equation}
	 \bm{c}_t\ = \ f_{att} (\bm{V}, \bm{h}_{t-1})
\end{equation}
In detail, the weights are calculated based on the correlation between visual features and the previous hidden state:
\begin{equation}
	 \bm{u}_{i,t}\ = \ \bm{w}_u \tanh (\bm{W}_{vu} \bm{x}_i + \bm{W}_{hu} \bm{h}_{t-1}) 
\end{equation}
\begin{equation}
\label{equ:alpha}
	 \alpha_t\ = \ \text{softmax} (\bm{u}_t)
\end{equation}
where each $\bm{x}_i$ of $\bm{V}$ represents the visual feature of $i$-th frame, $\bm{W}_{vu}$, $\bm{W}_{hu}$, and $\bm{w}_u$ are trainable parameters in $f_{att}$, $\alpha_t = \{\alpha_{1, t}, \alpha_{2, t}, \ldots, \alpha_{n, t}\} \in \mathbb{R}^{n}$ is a $n$-dimensional vector which sums to 1. The final attention feature $\bm{c}_t$ is generated by:
\begin{equation}
\label{equ:va-att}
	 \bm{c}_t\ = \ \sum_{i = 1}^{n} \alpha_{i,t} \bm{x}_i
\end{equation}

Take the correlation and the temporal coherence of the visual features extracted in the sequence frame as consideration, and we utilize an LSTM to construct the sequential coherence of the visual features extracted by the attention mechanism on each frame. Formally, for the $i$-th at step $t$:
\begin{equation}
	 \bm{h}_{i,t}^v\ = \ \text{LSTM2}_i (\alpha_i \bm{x}_i; \bm{h}_{i, t-1}^v)
\end{equation}
where $\bm{h}_{i, t-1}^v$ is the hidden state of $\text{LSTM2}_i$ at step $t$-1, and the parameters of $\text{LSTM2}_i$ on different frames are shared, which is conducive to constructing the correlation of visual features between different frames. These features are fused into the attention mechanism to guide the extraction of visual features at the next time step. Formally, the $\bm{c}_t$ is updated as:
\begin{equation}
	 \bm{c}_t\ = \ f_{att} (\bm{V}, \bm{h}_{t-1} \circ \bm{h}_{t-1}^v)
\end{equation}
\begin{equation}
	 \bm{h}_{t-1}^v\ = \ \sum_{i = 1}^{n} \bm{h}_{i, t-1}^v
\end{equation}
 % where the $\circ$ represents the concatenation operation.

 % \begin{equation}
 % \textbf{V}^{\prime} = \left[\begin{array}{c}\alpha_{1}\\\alpha_{2}\\\cdot\\\cdot\\\cdot\\\alpha_{m}\end{array}\right]\odot\left[\begin{array}{c}v_{1}\\v_{2}\\\cdot\\\cdot\\\cdot\\v_{m}\end{array}\right] = \left[\begin{array}{c}\alpha_{1}v_{1}\\\alpha_{2}v_{2}\\\cdot\\\cdot\\\cdot\\\alpha_{m}v_{m}\end{array}\right]
 % \end{equation}

\subsection{\textbf{Dual-stream Decoder}}
\label{sec:dual-stream}
 % 这个地方还要加一个总括性的表达, 就是设置了两条分支, 第一条分支为teacher-forcing, 第二条分支为semantic-guidance, 然后teacher-forcing分支用上标1表示, semantic-guidance用上标2表示
We design a dual-stream decoder to generate sentences, which includes the TF and SF stream, respectively. The TF stream is a traditional decoder, and the SF stream is designed to utilize the semantic information in the probability distribution over the words, which is beneficial to solving one-time ambiguity problems. To generate a text sequence, neural language model generate every token $y_t$ conditioned on the previous tokens in an auto-regressive manner:
\begin{equation}
\label{equ:teacher-forcing}
	 \log p_{\theta} (\bm{y} \mid \bm{x})\ = \ \sum_{t = 1}^{m} \log p_{\theta} (y_t \mid \bm{y}_{<t}, \bm{x})
\end{equation}
where $\theta$ are model parameters, and $\bm{y}_{<t}$ indicates all tokens before step $t$. 

\subsubsection{\textbf{Teacher-forcing Stream}}
Similar to the previous method, the TF scheme is considered in training. The word tokens $\bm{y}_{<t}^{\ast}$ from the ground-truth sequence are fed into~\eqref{equ:teacher-forcing} to generate the next token. Specifically, in video captioning, an LSTM is used to unroll the target information. At the $t$-th step, the target hidden state $h_t^{l_t}$ is given by:
\begin{equation}
\label{equ:tf-hidden}
	 \bm{h}_t^{l_t}\ = \ \text{LSTM3} ([\bm{W}_e{\bm{y}_{t-1}^{\ast}}, \bm{c}_t^{l_t}], \bm{h}_{t-1}^{l_t})
\end{equation}
where $W_e$ represents the word embedding, which maps the one-hot vector to word vector. $\bm{y}_{t-1}^{\ast}$ is the ground-truth word at step $t$-1, $\bm{c}_t^{l_t}$ is the visual feature extracted by the visual-aware attention mechanism mentioned in~\eqref{equ:va-att}. The probability distribution $\bm{p}_t^{l_t}$ over all the words in the target vocabulary is produced conditioned on the hidden state $\bm{h}_t^{l_t}$:
\begin{equation}
	 \label{tab:wp}
	 \bm{p}_t^{l_t}\ = \ \text{softmax} (\bm{W}_p \bm{h}_t^{l_t})
\end{equation}
where $\bm{W}_p$ is trainable parameter, $\bm{p}_t^{l_t}$ is a $D$-dimensional vector of vocabulary size.
 % which is used to map $\bm{h}_t^{l_t}$ to a dimension of one-hot vector so that each target word has one corresponding dimension in $\bm{P}_t^{l_t}$.

\subsubsection{\textbf{Self-forcing Stream}}
To take advantage of the complete semantic information embedded in $\bm{p}_t^{l_t}$, an SF stream parallel to the TF stream is designed. In detail, we feed the probability distributions obtained at the previous time step to the LSTM to promote the current token perceive richer and more specific information about the past:
\begin{equation}
	 \label{equ:sf-hidden}
	 \bm{h}_t^{l_s}\ = \ \text{LSTM4} ([\bm{W}_e \bm{p}_{t-1}^{l_s}, \bm{c}_t^{l_s}], \bm{h}_{t-1}^{l_s})
\end{equation}
\begin{equation}
	 \bm{p}_t^{l_s}\ = \ \text{softmax} (\bm{W}_p \bm{h}_t^{l_s})
\end{equation}
where the $\bm{h}_t^{l_s}$ is the hidden state of $\text{LSTM4}$ in SF stream, and the $\bm{c}_t^{l_s}$ is the visual feature by the visual-aware attention mechanism. Note that the SF stream shares the $\bm{W}_e$ and $\bm{W}_p$ parameters with the TF stream, which maps the words generated by the two-stream to the same semantic space.

 % We designed a dual-stream decoder to use the information in $P_{j}$. In detail, a teacher-forcing stream is used to learn the probability distributions of token supervised by the ground-truth. A modified student-forcing learning stream is used to exploit exploit the semantic information in P. which can also alleviate the problems of exposure bias caused by the discrepancy between training and testing. 
 % On the one hand, all the probability distributions obtained at the previous time step are used for the input of the current time step to make the current time step perceive richer and more specific information about the previous time step:

\subsection{\textbf{Training and Inference}}
\label{sec:mtl}
For full use of the semantic similarity between words, we train the whole network in an end-to-end manner by mixed training learning.
% the proposed model is jointly trained under the guidance of common teacher-forcing learning and proposed self-forcing learning.

\subsubsection{\textbf{Training}}
In training phase both the TF and SF LSTM units are executed for $m$ steps where $m$ refers to the length of ground-truth $\bm{Y}^{\prime}$. This process will generate two predicted sequences: $\bm{Y}^{t} = \{y_{1}^{t}, y_{2}^{t}, \ldots, y_{m}^{t}\}$ and $\bm{Y}^{s} = \{y_{1}^{s}, y_{2}^{s}, \ldots, y_m^{s}\}$, where $\bm{Y}^{t}$ corresponds to TF stream and $\bm{Y}^{s}$ for SF stream. The maximum likelihood estimation between the ground-truth and $\bm{p}_t^{l_t}$, $\bm{p}_t^{l_t}$ is used to training the TF stream and SF stream, respectively. The total loss is defined as:
\begin{equation}
	 \label{equ:train-loss}
	 \mathcal{L} (\theta)\ = \ \mathcal{L}_t + \lambda \mathcal{L}_{s}
\end{equation}
where $\mathcal{L}_t$ refers to the loss for the TF stream and $\mathcal{L}_{s}$ for the SF stream, and $\lambda$ is a hyper-parameter to balance the two terms. In detail, in the TF stream:
\begin{equation}
	 \mathcal{L}_t\ = \ -\frac{1}{m} \sum_{t = 1}^{m} \log (\bm{p}_t^{l_t} (y_t \mid \bm{y}_{1: t-1}^{\ast}))
\end{equation}
and in the SF stream:
 % \begin{equation}
 % \mathcal{L}_{s} = \delta \mathcal{L}_{s_{MLE}} + (1-\delta) \mathcal{L}_{s_{KL}}
 % \end{equation}
\begin{equation}
	 \mathcal{L}_{s}\ = \ -\frac{1}{m} \sum_{t = 1}^{m} \log (\bm{p}_t^{l_s} (y_t \mid \bm{p}_{1: t-1}^{l_s}))
\end{equation}

 % \begin{equation}
 % \mathcal{L}_{s_{KL}} = -\sum_{t = 1}^{m} \sum_{d \in \bm{D}} p_t^{{l_t}^{d}} \cdot \log \frac{p_t^{{l_s}^{d}}}{p_t^{{l_t}^{d}}}
 % \end{equation}
Note that the SF stream generates the current token based on the previous probability distributions $\bm{p}_{1: t-1}^{l_s}$.
 % and the $\mathcal{L}_{s_{KL}}$ is used to transfer the knowledge learned by teacher-forcing stream to self-forcing stream, and the $\delta$ is a dynamic value to control the flow from the TF stream to the SR stream, formally:

 % \begin{equation}
 % \delta = \frac{\mu}{\mu + \exp (e / \mu)}
 % \end{equation}
 % The function is strictly monotone decreasing, where $\mu$ is a hyper-parameter to control the strength of decreasing. At the beginning of training, as the TF stream is not well trained, maximum likelihood estimation is mainly used to train the SR stream; While at the end of training, the richer semantic features contained in the TF stream is utilize to train the SR stream through KL-Loss.
 % 关于超参, 训练的超参用控制重要性来表达, 测试的时候的超参用控制贡献来表达

\begin{figure}
	 \centering
	 \includegraphics[width = \columnwidth]{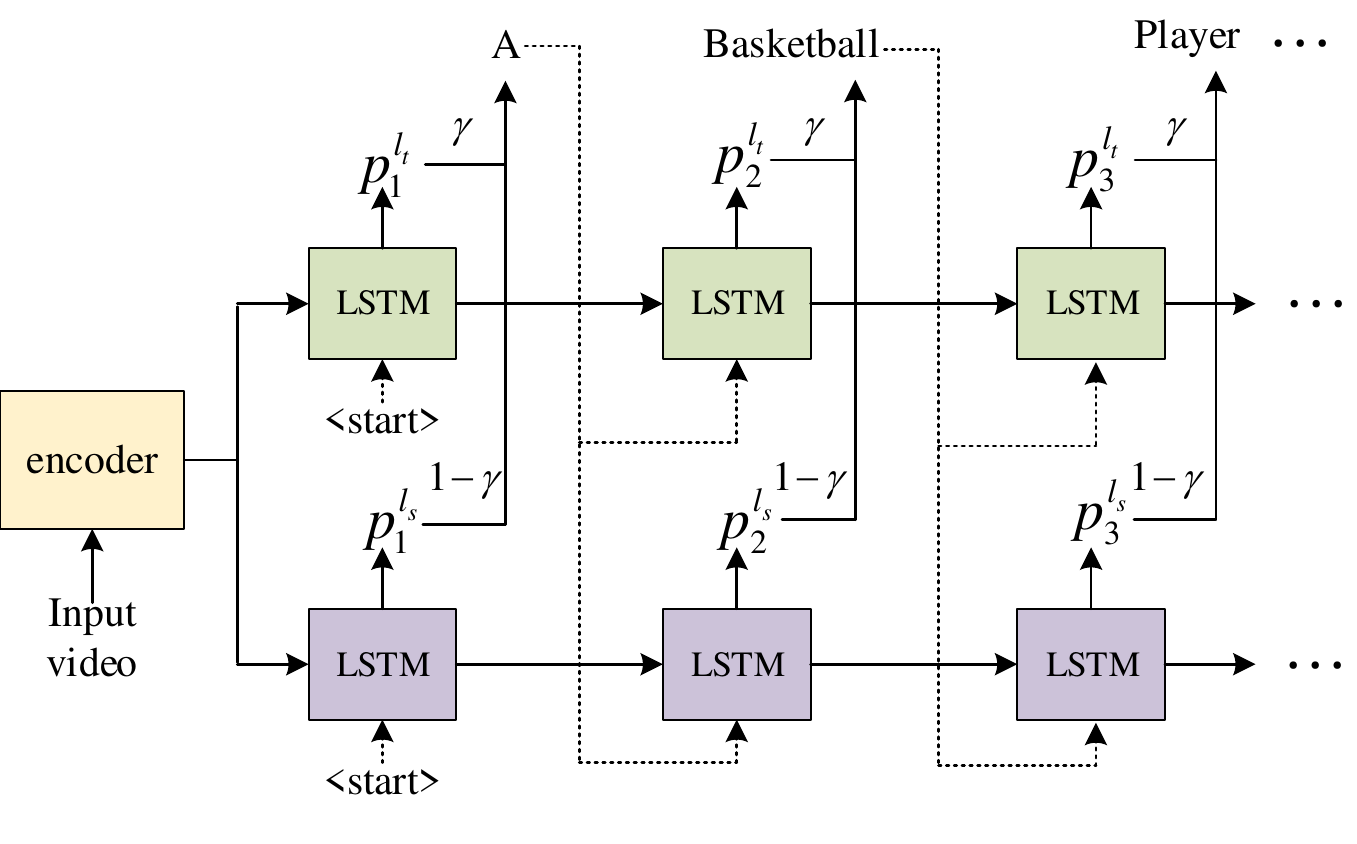}
	 \caption{\small The inference process for our VADD model. We extract information from the TF stream and SF stream in a certain proportion $\lambda$ to predict the current token at each step.}
	 \label{fig:inference}
	 %\vspace{-3mm}
\end{figure}

\subsubsection{\textbf{Inference}}
Figure~\ref{fig:inference} shows the inference process for our VADD methods. To utilize the information learned in the two stream, we combine the predicted probability of $p_t^{l_t}$ and $p_t^{l_s}$ together by:
\begin{equation}
\label{equ:testing-gamma}
	 p_t^{\prime}\ = \ \gamma p_t^{l_t} + (1-\gamma) p_t^{l_s}
\end{equation}
where $\gamma$ is used to control the proportion of information flow obtained from the two streams.

\section{\textbf{Experiments}}
In this section, we evaluate our proposed methods on two datasets: \textbf{MSVD} and \textbf{MSR-VTT}, via four popular metrics including BLEU-4, METEOR, ROUGE-L, and CIDEr, where the BLEU-4 metric mainly focuses on the fraction of $n$-grams between the ground-truth and the generated sentences, and the CIDEr metric is proposed for captioning tasks specifically and considered more consistent with human judgment. 
 % Our results are compared with state-of-art results, which demonstrate the effectiveness of our methods.

\subsection{\textbf{Datasets}}
\textbf{MSVD} contains 1,970 YouTube short video clips. Each video is annotated with multilingual sentences, but we experiment with the roughly 40 captions in English. Similar to the prior work~\cite{DBLP:conf/iccv/VenugopalanRDMD15}, we separate the dataset into 1,200 train, 100 validation, and 670 test videos.

\textbf{MSR-VTT} is another benchmark for video captioning which contains 10,000 open domain videos and each video is annotated with 20 English descriptions. There are 20 simple-defined categories, \textit{e.g.} music, sports, movie, \textit{etc}. We use the standard split in~\cite{DBLP:conf/cvpr/XuMYR16} for fair comparison which separates the dataset into 6,513 training, 497 validation, and 2,990 test videos.

\subsection{\textbf{Implementation Details}}
For the sentences in the benchmark datasets motioned above, we obtain a vocabulary by removing those rare words in training split with a threshold of two. We do minimum pre-processing to the annotated captions, \textit{i.e.} convert them into lower case and remove punctuation. We add the $\textless start \textgreater$ and $\textless end \textgreater$ at the beginning and end of each caption, respectively, and the words that are not contained in vocabulary are replaced with $\textless unk \textgreater$ token. We fix the length of sentences as 20, truncating those over-length sentences and adding $\textless pad \textgreater$ token at the end of under-length sentences.

For videos, we sample 50 frames for each video and use ResNet-152~\cite{DBLP:conf/cvpr/HeZRS16} as 2D CNN, ResNeXt-101~\cite{DBLP:conf/cvpr/XieGDTH17} as 3D CNN to extract appearance features and motion features respectively. ResNet-152 is trained on ILSVRC-2012-CLS image classification dataset~\cite{DBLP:journals/ijcv/RussakovskyDSKS15} and ResNeXt-101 is trianed on Kinetics action classification dataset~\cite{DBLP:journals/corr/KayCSZHVVGBNSZ17}.

Our model is optimized by Adam Optimizer~\cite{DBLP:journals/corr/KingmaB14}, the initial learning rate is set to 0.0001 and divided by 3 every 5 epochs. The hidden size of the LSTM is to 512 and 1, 024 for \textbf{MSVD} and \textbf{MSR-VTT} datasets, respectively. During testing, we use beam search with size 4 for the final caption generation. 

\subsection{\textbf{Performance Comparison}}
To evaluate the effectiveness of our methods, we compare our model with state-of-the-art models for video captioning on both \textbf{MSVD} and \textbf{MSR-VTT}.

\begin{table}
	 \centering
	 \caption{\small Performance comparisons on \textbf{MSR-VTT} datasets. \textbf{Bold} numbers are the best results.}
	 \footnotesize
	 \setlength{\tabcolsep}{3pt}
	 \begin{tabular}{lcccccc}
	 \toprule
	 Method & Venue & B-4 & M & R & C\\
	 \midrule
% 	 M3-IC~\cite{DBLP:conf/cvpr/Wang000T18} & CVPR'18 & 38.1 & 26.6 & - & -\\
	 ResNet~\cite{DBLP:conf/cvpr/Wang00018} & CVPR '18 & 39.1 & 26.6 & 59.3 & 42.7\\
	 SibNet~\cite{DBLP:conf/mm/LiuRY18} & ACM MM '18 & 40.9 & 27.5 & 60.2 & 47.5\\
% PickNet~\cite{DBLP:conf/eccv/ChenWZH18} \dag & CVPR'18 & 41.3 & 27.7 & 59.8 & 44.1\\
	 LG-DenseLSTM~\cite{DBLP:conf/mm/ZhuJ19} & ACM MM '19 & 38.1 & 26.6 & - & 42.8\\
	 TDConvED~\cite{DBLP:conf/aaai/ChenPLYCM19} & AAAI '19 & 39.5 & 27.5 & - & 42.8\\
	 VRE~\cite{DBLP:conf/mm/ShiCJG19} & ACM MM '19 & 39.0 & 26.9 & 60.0 & 44.3\\
	 MGSA~\cite{DBLP:conf/aaai/ChenJ19} & AAAI '19 & 39.9 & 26.3 & - & 45.0\\
	 OA-BTG~\cite{DBLP:conf/cvpr/ZhangP19a} & CVPR '19 & 41.4 & 28.2 & - & 46.9\\
	 MARN~\cite{DBLP:conf/cvpr/PeiZWKST19} & CVPR '19 & 40.4 & 28.1 & 60.7 & 47.1\\
	 GRU-EVE~\cite{DBLP:conf/cvpr/AafaqALGM19} & CVPR '19 & 38.3 & \textbf{28.4} & 60.7 & 48.1\\
	 POS + CG~\cite{DBLP:conf/iccv/Wang00JWL19} & ICCV '19 & 42.0 & 28.2 & 61.6 & 48.7\\
	 Two-stream~\cite{GaoL19} & TPAMI '20 & 39.7 & 27.0 & - & 42.1\\
	 VideoTRM~\cite{ChenC20} & ACM MM '20 & 38.8 & 27.0 & - & 44.7\\
	 BiLSTM-CG~\cite{DBLP:journals/npl/ChenZLLGZ20} & NPL '20 & 39.1 & 27.7 & 59.9 & 46.4\\
	 STGCN~\cite{PanC20} & CVPR '20 & 40.5 & 28.3 & 60.9 & 47.1\\
	 SAAT~\cite{DBLP:conf/cvpr/ZhengWT20} & CVPR '20 & 40.5 & 28.2 & 60.9 & 49.1\\
%PMI-CAP~\cite{DBLP:conf/eccv/ChenJLJ20} & ECCV'20 & 42.2 & \textbf{28.8} & - & 49.5\\
	 SGN~\cite{DBLP:conf/aaai/RyuK21} & AAAI '21 & 40.8 & 28.3 & 60.8 & 49.5\\
	 \midrule
	 Baseline & & 39.3 & 27.8 & 60.1 & 47.3\\
	 VA (ours) & & 41.0 & 28.1 & 60.7 & 47.8\\
	 DD (ours) & & 41.6 & 28.0 & 60.9 & 48.9\\
% 	 VATF (ours) & & 42.0 & 28.2 & 61.3 & 48.9\\
	 VADD (ours) & & \textbf{42.4} & 28.2 & \textbf{61.7} & \textbf{49.7}\\
%VAA-MTL \dag & & 41.4 & 28.4 & 62.2 & 53.0\\
	 \bottomrule
	 \label{tab:msr-vtt}
	 \end{tabular}
\end{table}

\begin{table}
	 \centering
	 \caption{\small Performance comparisons on \textbf{MSVD} dataset. \textbf{Bold} numbers are the best results.}
	 \footnotesize
	 \setlength{\tabcolsep}{3pt}
	 \begin{tabular}{lcccccc}
	 \toprule
	 Method & Venue & B-4 & M & R & C\\
	 \midrule
% Tube~\cite{DBLP:conf/ijcai/ZhaoLL18} & IJCAI'18 & 43.8 & 32.6 & 69.3 & 52.2\\
	 TSA-ED~\cite{DBLP:conf/cvpr/WuLCJL18} & CVPR '18 & 51.7 & 34.0 & - & 74.9\\
	 PickNet~\cite{DBLP:conf/eccv/ChenWZH18} & CVPR '18 & 46.1 & 33.1 & 69.2 & 76.0\\
	 ResNet~\cite{DBLP:conf/cvpr/Wang00018} & CVPR '18 & 52.3 & 34.1 & 69.8 & 80.3\\
    %  M3-IC~\cite{DBLP:conf/cvpr/Wang000T18} & CVPR'18 & 52.8 & 33.3 & - & -\\
	 SibNet~\cite{DBLP:conf/mm/LiuRY18} & ACM MM '18 & 54.2 & 34.8 & 71.7 & 88.2\\
	 GRU-EVE~\cite{DBLP:conf/cvpr/AafaqALGM19} & CVPR '19 & 47.9 & 35.0 & 71.5 & 78.1\\
	 LG-DenseLSTM~\cite{DBLP:conf/mm/ZhuJ19} & ACM MM '19 & 50.4 & 32.9 & 69.9 & 72.6\\
	 VRE~\cite{DBLP:conf/mm/ShiCJG19} & ACM MM '19 & 51.7 & 34.3 & 71.9 & 86.7\\
	 POS + CG~\cite{DBLP:conf/iccv/Wang00JWL19} & ICCV '19 & 52.5 & 34.1 & 71.3 & 88.7\\
	 POS + VCT~\cite{DBLP:conf/iccv/HouWZLJ19} & ICCV '19 & 52.8 & \textbf{36.1} & 71.8 & 87.8\\
	 FCVC-CF~\cite{DBLP:conf/aaai/FangZJZWZF19} & AAAI '19 & 53.1 & 34.8 & 71.8 & 79.8\\
	 TDConvED~\cite{DBLP:conf/aaai/ChenPLYCM19} & AAAI '19 & 53.3 & 33.8 & - & 76.4\\
	 MGSA~\cite{DBLP:conf/aaai/ChenJ19} & AAAI '19 & 53.4 & 35.0 & - & 86.7\\
% OA-BTG~\cite{DBLP:conf/cvpr/ZhangP19a} & CVPR'19 & 56.9 & 36.2 & - & 90.6\\
	 Two-stream~\cite{GaoL19} & TPAMI '20 & \textbf{54.3} & 33.5 & - & 72.8\\
	 SAAT~\cite{DBLP:conf/cvpr/ZhengWT20} & CVPR '20 & 46.5 & 33.5 & 69.4 & 81.0\\
	 BiLSTM-CG~\cite{DBLP:journals/npl/ChenZLLGZ20} & NPL '20 & 53.3 & 35.2 & 71.6 & 84.1\\
	 SBAT~\cite{DBLP:conf/ijcai/JinHCLZ20} & IJCAI '20 & 53.1 & 35.3 & 72.3 & 89.5\\
	 \midrule 
	 Baseline & & 49.1 & 34.6 & 71.8 & 87.9\\ 
	 VA (ours) & & 51.8 & 34.8 & 72.2 & 88.8\\
	 DD (ours) & & 52.8 & 34.8 & \textbf{72.7} & 89.0\\
% 	 VATF (ours) & & 50.2 & 34.1 & 71.9 & 90.9\\
	 VADD (ours) & & 51.5 & 34.8 & 72.1 & \textbf{91.5}\\
	 \bottomrule
	 \label{tab:msvd}
	 \end{tabular}
\end{table}

\subsubsection{\textbf{Ablation Study}}
To demonstrate the effectiveness of the proposed components, we separate our model for ablation study on \textbf{MSVD} and \textbf{MSR-VTT}. There are several different settings: a) Baseline: the traditional attention mechanism is utilized to generate sentences under the way of TF learning during training; b) VA refers to the visual-aware attention mechanism is applied to construct temporal coherence and correlation of visual features obtained through the attention mechanism between different frames; c) DD refers to the dual-stream decoder (\textit{e.g.} TF stream and SF stream) are applied to take advantage of the semantic information of generated words; d) VADD refers that the dual-stream decoder is utilized, and the visual-aware attention mechanism is attached to both TF stream and SF stream.
% 4) VATF means the method generate sentence by the dual-stream decoder, and in the TF stream, the visual-aware attention mechanism is utilized to extract visual features; 

The ablation experimental results of aforementioned models with different components are shown in Table~\ref{tab:msr-vtt} and Table~\ref{tab:msvd}. The experimental results is gradually increasing in baseline, VA, and VADD. Compared with the baseline, we can observe a improvement on all the evaluation metrics especially in BLEU and CIDEr for our proposed method. Specifically, on MSR-VTT dataset, our proposed method (VADD) improves BLEU and CIDEr by 3.1$\%$ and 2.4$\%$, respectively. On MSVD dataset, our method imporves them by 2.4$\%$ and 3.6$\%$. 
We observe that compared with combined visual-aware attention mechanism (VADD), the DD has a superior results on \textbf{MSVD} datasets for the BLEU-4 metric.
We analyze \textbf{MSVD} datasets contain far fewer video sentence pairs than \textbf{MSR-VTT} datasets, and most of the annotated captions are short sentences on \textbf{MSVD} datasets, increasing the challenge of training the visual-aware attention mechanism and weakens the effectiveness of its.
% In detail, the comparison between baseline and VA, DD confirms the effectiveness of the visual-aware attention mechanism and dual-stream decoder, respectively. The gap between the VATF and the DD verifies the effectiveness of the visual-aware attention mechanism combined with the TF stream.
% Correspondingly, VADD is superior to VATF, which verifies the effectiveness of our visual-aware attention mechanism on the SF stream. 
% We observe that compared with combined visual-aware attention mechanism (\textit{e.g.} VADD and VATF, the DD has a superior results on \textbf{MSVD} datasets for the BLEU-4 metric. We analyze that most of the annotated captions are short sentences on \textbf{MSVD} datasets, which weakens the effectiveness of the visual-aware attention mechanism. In addition, the \textbf{MSVD} datasets contain far fewer video sentence pairs than the \textbf{MSR-VTT} datasets, and this will increase the challenge of training the visual-aware attention mechanism.

\subsubsection{\textbf{Comparison with the State-of-the-arts}}
For \textbf{MSR-VTT} dataset, we use BLEU-4, METEOR, Rouge, and CIDEr to evaluate the generated sentences. As shown in Table~\ref{tab:msr-vtt}, the quantitative results across most metrics indicate that the proposed VADD method outperforms the traditional ones. And our methods are also comparable to those compared ones.

Our method is 1.6\% and 0.9\% higher than the recent method SGN~\cite{DBLP:conf/aaai/RyuK21}, on BLEU-4 and Rouge metrics, and has considerable effects on METEOR and CIDEr.
compared with the performances GRU-EVE~\cite{DBLP:conf/iccv/Wang00JWL19}, OA-BTG~\cite{DBLP:conf/cvpr/ZhangP19a}, SAAT~\cite{DBLP:conf/cvpr/ZhengWT20} based on the object appearance features (Detector), our method (VADD) significantly outperforms all above of them even without the object appearance features (Detector). 
% It also can be observed that VADD surpasses VA-TF a little bit in MSR-VTT. One possible reason for this phenomenon is that, when we use Attention-Aware mechanism to fuse dynamic visual feature of each frame, both dynamic relationship of the visual features extracted in each frame and the latest generated hidden states are used as queries to modulate all hidden states and the importance of them are the same.
% To evaluate the importance of different visual features, we also report the results of the proposed methods based on the object appearance features (Detector), RGB features, and motion features.
%Compared with the performances SAAT~\cite{DBLP:conf/cvpr/ZhengWT20} based on the object appearance features (Detector), our method VADD can be observed that the performances are much improved. 
Our method captures the critical details of the video frame and make full use of the semantic information of previous generated words, so our model achieves 49.7 on CIDEr, which makes an improvement over ResNet~\cite{DBLP:conf/cvpr/Wang00018} by 7.0\% and the most state-of-the-art method STGCN~\cite{PanC20} by a margin of 2.6\%. Our method is slightly lower than GRU-EVE~\cite{DBLP:conf/cvpr/AafaqALGM19} on the METEOR metric, which can be attributed to the GRU-EVE~\cite{DBLP:conf/cvpr/AafaqALGM19} method using additional dictionaries to provide semantic information.

Table~\ref{tab:msvd} shows the results of different methods on \textbf{MSVD} dataset. Compared with the recent video captioning methods, our method achieves the best performance in CIDEr and ROUGE\_L. Our CIDEr score is significantly higher than SBAT~\cite{DBLP:conf/ijcai/JinHCLZ20}, SAAT~\cite{DBLP:conf/cvpr/ZhengWT20}, and Two-Stream~\cite{GaoL19} by a large margin of 2\%, 10.5\%, and 18.7\%, which confirms that our model can generate sentences that are more semantically similar to the reference sentence. Our model does not achieve very prominent effects on the BLEU-4 and METEOR metrics on \textbf{MSVD} dataset. The BLEU-4 mainly focuses on the fraction of $n$-grams between the ground-truth and the generated sentences. Considering that our model has achieved high scores on the CIDEr metric, we have reason to assume that our model can generate sentences that have the same semantics as the reference sentence with different expressions.

\begin{table}
	 \centering
	 \caption{\small Performances evaluation of our methods with different $\lambda$ on \textbf{MSR-VTT} dataset where $\lambda$ is the trade-off parameter in~\eqref{equ:train-loss}. \textbf{Bold} numbers are the best results.}
	 \footnotesize
	 \label{tab:lambda_mtl}
	 \begin{tabular}{lccccc}
	 \toprule
	 Method & $\lambda$ & B-4 & M & R & C\\
	 \midrule
	 & 0.4 & 41.1 & 27.8 & 60.5 & 48.1\\ 
	 & 0.6 & 40.7 & 28.0 & \textbf{60.9} & 48.1\\
	 DD & 0.8 & \textbf{41.6} & 28.0 & \textbf{60.9} & \textbf{48.9}\\
	 & 1.0 & 41.1 & 28.0 & \textbf{60.9} & 48.2\\
	 & 1.2 & 41.4 & \textbf{28.2} & \textbf{60.9} & 48.6\\
 % 1.4 & 41.4 & 28.1 & 60.9 & 48.8\\
	 \midrule
	 & 0.4 & 41.5 & 28.1 & 61.4 & 48.5\\
	 & 0.6 & 41.4 & 28.0 & 61.1 & 49.1\\
	 VADD & 0.8 & \textbf{42.4} & \textbf{28.2} & \textbf{61.7} & \textbf{49.7}\\
	 & 1.0 & 41.8 & 27.9 & 61.6 & 48.8\\
	 & 1.2 & 41.6 & 27.9 & 61.2 & 49.0\\
 % 1.4 & 41.4 & 27.9 & 61.1 & 48.4\\ 
	 \bottomrule
	 \end{tabular}
\end{table}

\subsection{\textbf{Study on Trade-off Parameter}}
\subsubsection{\textbf{Trade-off Parameter of $\lambda$ in Training}}
In our proposed DD and VADD methods, we combine two generated sequences with a trade-off parameter $\lambda$ in~\eqref{equ:train-loss} to control the importance of the TF stream and SF stream during training. The results in Table~\ref{tab:lambda_mtl} show that both DD and VADD achieve their best performances at $\lambda$ = 0.8. This is reasonable because if $\lambda$ is too small, it will be tough for SF stream to model the semantic information in the probability distribution of the previous token; If $\lambda$ is too large, it will interfere with the TF stream during training. We find that the loss function of SF stream is greater than that of TF stream. Therefore, $\lambda$ = 0.8 plays a role in balancing the importance of the TF stream and the SF stream.

\begin{figure}
	 \centering
	 \includegraphics[width = \columnwidth]{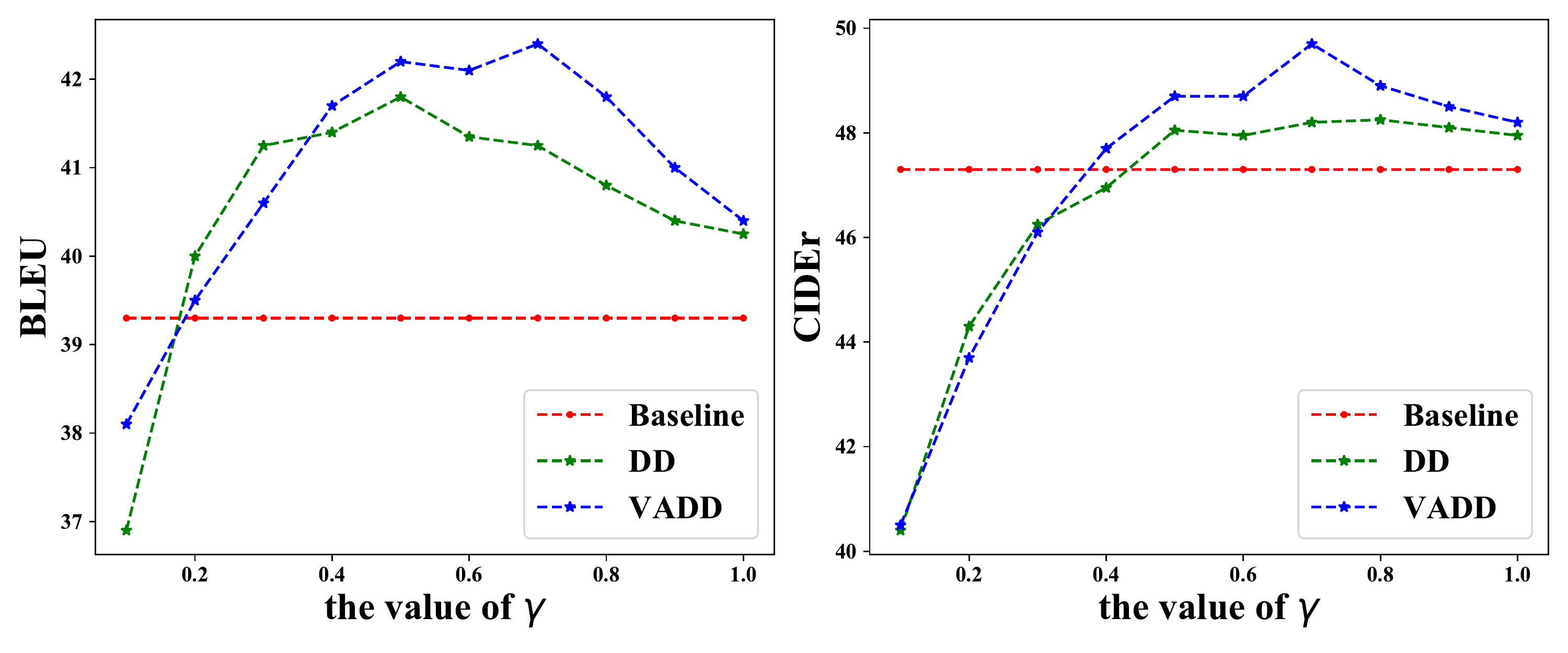}
	 \caption{\small Results on \textbf{MSR-VTT} with different $\gamma$.}
	 \label{fig:test_ratio}
	 %\vspace{-3mm}	 
\end{figure}

\begin{figure}
	 \centering
	 \includegraphics[width = \columnwidth]{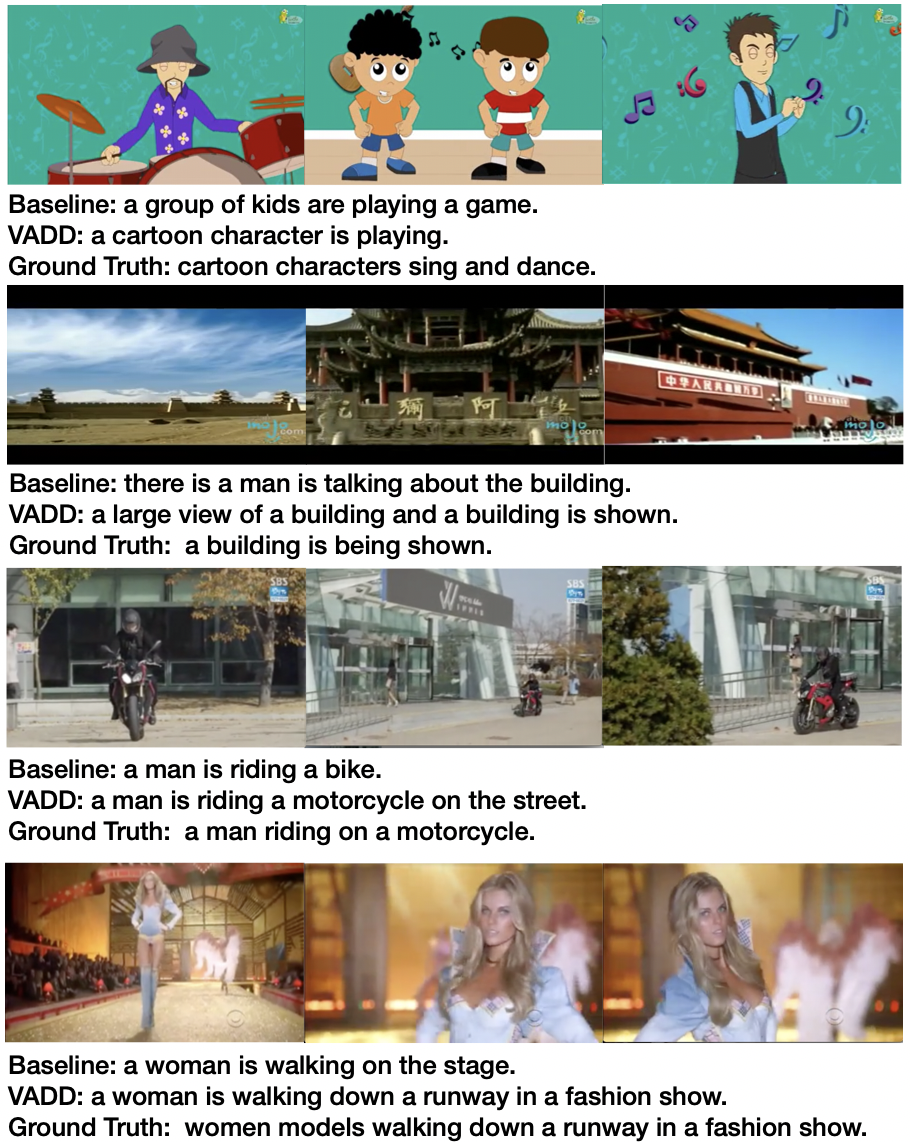}
	 \caption{\small The generated captioning of Baseline, our method (VADD), and the Ground Truth on \textbf{MSR-VTT} dataset.}
	 \label{fig:qualitative}
	 %\vspace{-3mm}
\end{figure}

\subsubsection{\textbf{Trade-off Parameter of $\gamma$ in Testing}}
To investigate the effect of $\gamma$ in~\eqref{equ:testing-gamma} on model accuracy, we conduct a series of experiments with different $\gamma$ on \textbf{MSR-VTT} datasets during the testing phase. In order to enhance the reliability of the algorithm, we take the average of two experiments. The results are shown in Figure~\ref{fig:test_ratio}, the model accuracy decreases gradually, no matter when the $\gamma$ is too big (\textit{i.e.}, the predicted tokens is depending on the TF stream) or too small (\textit{i.e.}, the predicted tokens is depending on the SF stream). We analyze that the SF stream contains rich semantic information in sentence generation, but it is not easy to train itself without ground-truth word guidance. In addition, we observe that the result is still better than the baseline at the $\gamma = 1.0$ in the DD model. This means that the SF stream is merely used during training, and the information in it is abandoned during testing, which can be attributed to the mutual learning between the TF stream and the SF stream through the shared parameters $\bm{W}_e$ and $\bm{W}_p$ in~\eqref{equ:tf-hidden} and~\eqref{tab:wp}. When $\gamma$ is relatively small, the effect of the DD model is better than the VADD model at certain values. We argue that although the visual-aware attention mechanism has a disturbance to the SF stream, the combination of the TF stream and the SF stream is not a linear superposition of the two. Due to the shared parameters $\bm{W}_e$ and $\bm{W}_p$, the application of the visual-aware attention mechanism to the SF stream will also indirectly affect the TF stream, which results in a negative effect on the single SF stream, while has a positive effect on the combination of the SF stream and the TF stream. Therefore, when $\gamma$ is relatively small, the SF stream plays a major role. As $\gamma$ increases, the combination of the SF stream and TF stream play a positive effect.
%前面这句话改一下

% \begin{figure*}
% \centering
% \includegraphics[width = \textwidth]{caption.png}
% \caption{\small The generated captioning of ground-truth, baseline, and our method on MSR-VTT dataset.}
% \label{fig:qualitative}
% \end{figure*}

\subsection{\textbf{Qualitative Results}}
Figure~\ref{fig:qualitative} shows cases of a few sentences generated by different methods and human-annotated ground-truth sentences. From these exemplar results, it is easy to see that all of these automatic methods can generate somewhat relevant sentences, while our proposed VADD can predict more relevant keywords, and generate a more accurate and coherent description. The baseline method have a weaker ability of recognition than our methods. \textit{e.g.} “a group of kids" in the first example in the baseline, while our method can integrate contextual information to generate words “a cartoon characters". Since there is no dynamic variational relationship of each frame, the baseline model tends to result in incorrect collocation, \textit{e.g.} “talking about the building” in the second example. Contrarily, VADD can generate additional contextual coherence \textit{e.g.} “a building is shown”. We can conclude that our proposed VADD can extract pivotal features dynamically and the semantic information in the probability distribution over the words from SF stream. In the third example, by enhancing the coherence between visions, our method can generate the correct description “motorcycle" while the baseline model generates the wrong word “bike". In the fourth example, our method generates high-level semantic words “fashion show" by combining contextual semantic information, while baseline can only generate the descriptive phrase “walking on the stage". All these examples once again demonstrate that our proposed method can capture more relevant visual information and richer semantic information.

%These results validate that our proposed VADD has a more vital ability to recognize contextual information.
%Looking at the third example, when our proposed VA model constructs the temporal coherence and dynamic relationship of the visual features extracted in each frame, the caption will lead to a relevant motion concept "makes a slam dunk". After combining VA and DD methods, it generates a better sentence that is more similar to the ground-truth. Therefore, these examples demonstrate that our method makes a significant improvement of the video captioning model again. We can conclude that our proposed VADD can extract pivotal features dynamically.

\section{\textbf{CONCLUSION}}
In this paper, a video captioning model named VADD is proposed to enhance visual-aware attention and dual-stream decoder. We design the visual-aware attention mechanism to utilize the correlation and the temporal coherence of the visual features extracted in the sequence frame. Our visual-aware attention method takes attention value from the previous word and dynamic visual feature into the input of the current attention module. Moreover, the TF stream is a traditional text training strategy. We exploit the SF stream to utilize the semantic information in the probability distribution over the words. In the training and inference phase, the two training learning strategies are combined to predict the current word. On two commonly used datasets, the experimental results demonstrate the effectiveness of the proposed approaches.

\bibliographystyle{IEEEtran}
\bibliography{VADD}

% \bibitem{b1} G. Eason, B. Noble, and I. N. Sneddon, ``On certain integrals of Lipschitz-Hankel type involving products of Bessel functions, '' Phil. Trans. Roy. Soc. London, vol. A247, pp. 529--551, April 1955.
% \bibitem{b2} J. Clerk Maxwell, A Treatise on Electricity and Magnetism, 3rd ed., vol. 2. Oxford: Clarendon, 1892, pp.68--73.
% \bibitem{b3} I. S. Jacobs and C. P. Bean, ``Fine particles, thin films and exchange anisotropy, '' in Magnetism, vol. III, G. T. Rado and H. Suhl, Eds. New York: Academic, 1963, pp. 271--350.
% \bibitem{b4} K. Elissa, ``Title of paper if known, '' unpublished.
% \bibitem{b5} R. Nicole, ``Title of paper with only first word capitalized, '' J. Name Stand. Abbrev., in press.
% \bibitem{b6} Y. Yorozu, M. Hirano, K. Oka, and Y. Tagawa, ``Electron spectroscopy studies on magneto-optical media and plastic substrate interface, '' IEEE Transl. J. Magn. Japan, vol. 2, pp. 740--741, August 1987 [Digests 9th Annual Conf. Magnetics Japan, p. 301, 1982].
% \bibitem{b7} M. Young, The Technical Writer's Handbook. Mill Valley, CA: University Science, 1989.
% \end{thebibliography}

\end{document}